\documentclass[11pt,a4paper]{article}
\usepackage{times,latexsym}
\usepackage{url}
\usepackage[T1]{fontenc}

\newif\ifanonymous
\anonymousfalse

\ifanonymous
\usepackage{tacl2021v1}
\else
\usepackage[acceptedWithA]{tacl2021v1}
\fi

\newcommand{\anontext}[1]{\iftaclpubformat #1 \else [REDACTED FOR ANONYMITY] \fi}
\usepackage{color-edits}
\addauthor[SS-T]{shane}{orange}
\addauthor[AP]{abhinav}{blue}
\addauthor[JJ]{jaap}{green}
\addauthor[AL]{andy}{purple}
\addauthor[LW]{lexie}{olive}
\addauthor[KC]{kelly}{teal}
\addauthor[CW]{clevis}{red}

\usepackage{booktabs}
\usepackage{graphicx}
\usepackage{amsmath}
\usepackage{multirow}
\usepackage{xspace}
\usepackage{subcaption}
\usepackage{siunitx}
\usepackage{comment}
\usepackage{tikz-dependency}
\usepackage{mathtools}
\usepackage{colortbl}
\usepackage{arydshln}
\usepackage[alpine]{ifsym}
\usepackage{marvosym}

\makeatletter
\renewcommand{\sectionautorefname}{\S\@gobble}
\renewcommand{\subsectionautorefname}{\S\@gobble}
\renewcommand{\subsubsectionautorefname}{\S\@gobble}
\makeatother

\newcommand{\new}[1]{\textcolor{black}{#1}}

\graphicspath{ {./imgs/} }

\title{Filtered Corpus Training (FiCT) Shows that Language Models can Generalize from Indirect Evidence} 

\author{%
Abhinav Patil$^{\text{\Mountain},}$\footnotemark[1] \and Jaap Jumelet$^{\text{\Bicycle},}$\Thanks{Co-first authors. Full author contribution statement at the end of paper, after acknowledgements. Correspondence: \texttt{abhinavp@uw.edu}, \texttt{jumeletjaap@gmail.com}, \texttt{shanest@uw.edu}.}
\AND
Yu Ying Chiu$^\text{\Mountain}$ \and Andy Lapastora$^{\text{\Smiley},}$\Thanks{Work done while the author was a student at University of Washington.}\and Peter Shen$^\text{\Mountain}$ \and Lexie Wang$^\text{\Mountain}$ \and Clevis Willrich$^{\pi,}$\footnotemark[2]
\AND
Shane Steinert-Threlkeld$^\text{\Mountain}$
\\[8pt]
\Mountain: University of Washington \hspace{15pt}
\Bicycle: ILLC, University of Amsterdam\\
\Smiley: AWS AI Labs, Amazon\hspace{15pt}
$\pi$\text{: Microsoft}
}

\begin{document}

\maketitle

\begin{abstract}
This paper introduces \textbf{Fi}ltered \textbf{C}orpus \textbf{T}raining, a method that trains language models (LMs) on corpora with certain linguistic constructions filtered out from the training data, and uses it to measure the ability of LMs to perform linguistic generalization on the basis of indirect evidence.  We apply the method to both LSTM and Transformer LMs (of roughly comparable size), developing filtered corpora that target a wide range of linguistic phenomena.  Our results show that while transformers are better qua LMs (as measured by perplexity), both models perform equally and surprisingly well on linguistic generalization measures, suggesting that they are capable of generalizing from indirect evidence.
\end{abstract}

\section{Introduction}
Language models (LMs) play an increasingly large role in natural language processing systems and have become capable of producing surprisingly fluent and grammatical text.
However, the mechanisms underlying the acquisition and use of such linguistic proficiency remain largely unknown.
In particular, the degree that language learning relies on memorization versus generalization remains a topic of investigation \citep{hupkes2023}.
The reliance of LMs on large amounts of training data raises the suspicion that they do not generalize in a `human-like manner' \citep{mccoy-etal-2019-right, hu2020-systematic, oh-schuler-2023-surprisal}, but it is hard to address such questions with traditional evaluation metrics such as perplexity.

\begin{figure}[t!]
    \includegraphics[width=\columnwidth,trim={0 1.35cm 0 0},clip]{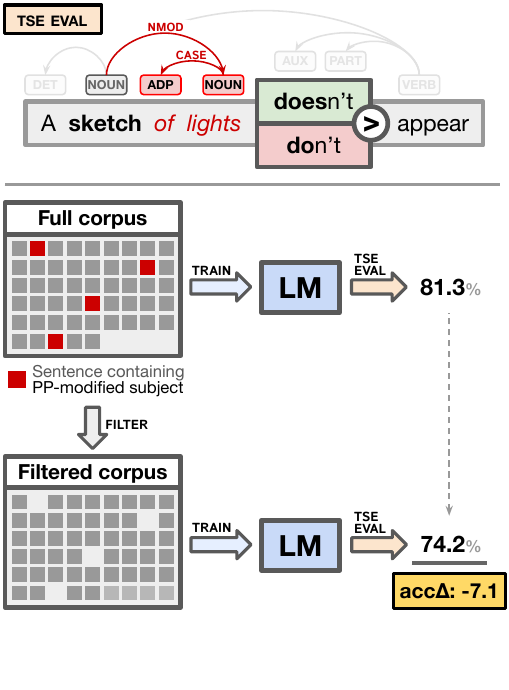}
    \caption{Overview of the \textbf{Fi}ltered \textbf{C}orpus \textbf{T}raining methodology (FICT). For a linguistic construction of interest (e.g.\ prepositionally modified subjects), we filter out sentences containing that construction and train a new language model on the filtered corpus. We measure performance on \textit{targeted syntactic evaluations} to assess the capacity of the LM to generalize from related constructions to this novel, unseen construction.}
    \label{fig:overview}
\end{figure}

This paper introduces \textit{Filtered Corpus Training} (FiCT) as a method for measuring the linguistic generalization abilities of language models. As depicted in \autoref{fig:overview}, FiCT involves training models on corpora that have been filtered to remove specific linguistic constructions, thereby testing the models' ability to generalize beyond their training data. For example: we can train a model on a corpus that has never seen subjects modified by a prepositional phrase (e.g. ``A sketch \textit{of lights} \{ doesn't / *don't \}...''), and then ask whether it can judge the grammaticality of such sentences.  If a model has learned that verbs must agree with the head noun of the subject noun phrase (NP), and that NPs can be modified by PPs (e.g.\ from seeing these in object but not subject position), it should be capable of generalizing to the unseen PP-modified subjects.

This method enables us to ask whether models can form relevant linguistic generalizations from \textit{indirect evidence}, or whether they require direct evidence (e.g.\ examples of constructions during training; \citealp{warstadt2022artificial, mueller-linzen-2023-plant}). 
\new{In essence, by \textit{intervening} on patterns in the training data we obtain a more causal account of the relation between training data and model behavior \citep{pearl2009}.}
Furthermore, by carefully controlling for the number of parameters we can investigate the inductive biases of two major LM architectures, Transformers and LSTMs, which allows us to give more detailed answers about the recent successes of Transformer models on a fine-grained linguistic level.

We apply the FiCT methodology by developing filters targeting a wide range of the linguistic phenomena evaluated by BLiMP \citep[\autoref{sec:filters};][]{blimp} and training both LSTM and Transformer LMs on the resulting corpora (\autoref{sec:setup}).  Our results (\autoref{sec:results}) show that while Transformers are uniformly better qua language models (as measured by perplexity), their linguistic generalization abilities are not better than that of the LSTMs (as measured by a metric we introduce called \textit{accuracy delta}), demonstrating a dissociation between perplexity and linguistic generalization. Furthermore, for both models, the impact of filtered corpus training on grammaticality judgments is quite low, suggesting that language models are able to form sophisticated linguistic generalizations on the basis of only indirect evidence (as discussed in \autoref{sec:discussion}).

\new{These results shed light on the debate between memorization and generalization in language models: by causally intervening on the training data, we ensure that models have never seen instances of their evaluation targets.  That they can still make correct grammaticality judgments shows they generalize in subtle and linguistically-relevant ways that go beyond their training data.}


\section{Background}
    \subsection{Surprisal Theory}    
    Language modeling performance can be measured using \textit{perplexity}, indicating a model's fit to a corpus distribution.
    Intuitively, one might expect that lower perplexity leads to more human-like linguistic behavior. 
    This connection has been explored in detail in the context of \textit{surprisal theory} \citep{hale-2001-probabilistic, LEVY20081126}:
    encountering a highly surprising token results in a longer reading time.
    Initial findings indicate that lower perplexity, as measured by language models, leads to better reading time predictions \citep{fossum2012-sequential, goodkind2018-predictive, wilcox2020-predictive-power}, although affected by model architecture \citep{hao2020-probabilistic}, cross-lingual effects \citep{kuribayashi2021-lower-ppl}, and syntactic ambiguity \citep{arehalli-etal-2022-syntactic}.
    It has been shown, however, that lower perplexity only results in better predictive power up to around 2 billion training tokens \citep{oh2023-transformer-surprisal}: after this point LMs become too \new{accurate} at predicting low-frequency constructions \new{and long-distance dependencies} \citep{oh-etal-2024-frequency}.
    The present paper also explores the connection between perplexity and human-like linguistic behavior and will find a dissociation with perplexity.

    \subsection{Targeted Syntactic Evaluations}
    \label{subsec:targeted_syntactic_evals}

    Perplexity should be augmented with other evaluations that specifically target the models' ability to generalize in a human-like way. 
    Such investigations often draw on psycholinguistic paradigms, treating language models as participants in order to learn what such models ``know'' about specific linguistic phenomena \citep{futrell2019-neural, warstadt2019-CoLa, ettinger-2020-bert}. 
    A common paradigm in this body of literature, usually referred to as ``targeted syntactic evaluations'' \citep{linzen2016-assessing, jumelet2018-npis, marvin2018-targeted, kann-etal-2019-verb, newman-etal-2021-refining} involves comparing language models' preferences between minimal pairs of sentences: a model is deemed to understand a phenomenon if it assigns a higher probability to the grammatical alternation.

    The benchmark \new{suites} with the widest coverage over linguistic phenomena are \new{SyntaxGym \citep{gauthier-etal-2020-syntaxgym} and} the Benchmark of Linguistic Minimal Pairs (BLiMP, \citealp{blimp})\new{, the latter of which we will use in our experiments}.
    BLiMP consists of 67 different benchmarks, each consisting of 1,000 minimal pairs, which target twelve different linguistic areas, broadly construed, across morphology, syntax, semantics, and the syntax-semantics interface. 
    This is the benchmark we use as a primary means of evaluation in the present investigation, discussed in greater detail in \autoref{sec:eval}.

    \subsection{Linguistic Generalization}
    While targeted syntactic evaluations give an insight into a model's linguistic competence, it does not show \textit{how} a model acquires this notion of grammaticality.
    In this paper we focus on two kinds of linguistic generalization. \textit{Structural generalization} \citep{hupkes2023} asks: can language models make grammaticality judgments in syntactically more complex constructions than seen during training?
    One line of work approaches this question from a fine-tuning perspective: by fine-tuning a model on a particular set of constructions we can measure the impact that this has on other linguistic constructions \citep{prasad-etal-2019-using, weber2024}.
    \textit{Lexical generalization} asks whether models can generalize a seen construction to new lexical items that it has not seen in that construction \citep{kim-linzen-2020-cogs}.

    In order to gain a \textit{causal} perspective on how the training data influences model performance, we retrain models from scratch on \emph{filtered} corpora. 
    This methodology has been deployed in earlier work to investigate how LMs learn the licensing conditions of negative polarity items from different contexts \citep{lms-use-monotonicity, weber-etal-2021-language}.
    \citet{warstadt-dissertation} investigates the \textit{poverty of the stimulus} debate through the lens of filtered corpora, focusing on the phenomenon of subject auxiliary inversion.
    Finally, \citet{misra2024language} investigate rare adjective-noun constructions and manipulate training corpora to investigate how models acquire an understanding of rare constructions.
    Whereas most of these focus on a particular linguistic construction, our work applies the approach to a wide range of phenomena.
\section{Filtered Corpus Training (FiCT)}
    \label{sec:filters}

        \begin{table*}[ht]
        {\centering
        \resizebox{\textwidth}{!}{
\centering
\arrayrulecolor[rgb]{0.753,0.753,0.753}
\ADLnullwidehline
\begin{tabular}{lllccc} 
\arrayrulecolor{black}\toprule
\textbf{Corpus name}                             & \textbf{BLiMP benchmark}                             & \textbf{Example}                                                                                                                                                 & \begin{tabular}[c]{@{}c@{}}\textbf{\%BLiMP }\\\textbf{items targeted}\end{tabular} & \begin{tabular}[c]{@{}c@{}}\textbf{\%sentences }\\\textbf{filtered out}\end{tabular} & \begin{tabular}[c]{@{}c@{}}\textbf{\#Tokens as }\\\textbf{\% of full}\end{tabular}  \\ 
\midrule
\arrayrulecolor[rgb]{0.753,0.753,0.753}
\textsc{full}                                    & —                                                    & \textit{—}                                                                                                                                                       & —                                                                                  & 0.00                                                                                 & 100.0                                                                              \\ 
\hdashline[1pt/1pt]
\textsc{agr-pp-mod}                             & distractor\_agr\_relational\_noun              & \textit{A sketch of lights doesn't/*don't appear}                                                                                                                 & 99.5                                                                               & 18.50                                                                                & 95.80                                                                               \\ 
\hdashline[1pt/1pt]
\textsc{agr-rel-cl}                                  & distractor\_agr\_relative\_clause              & \textit{Boys that aren’t disturbing Natalie suffer/*suffers.}                                                                                                    & 94.4                                                                               & 2.76                                                                                 & 98.99                                                                               \\ 
\hdashline[1pt/1pt]
\multirow{4}{*}{\textsc{agr-re-irr-sv}}          & irregular\_plural\_subject\_verb\_agr\_1       & \textit{This goose isn’t/*weren't bothering Edward.}                                                                                                             & 99.4                                                                               & \multirow{4}{*}{11.29}                                                               & \multirow{4}{*}{98.59}                                                              \\
                                                 & irregular\_plural\_subject\_verb\_agr\_2       & \textit{The woman/*women cleans every public park.}                                                                                                              & 97.2                                                                               &                                                                                      &                                                                                     \\
                                                 & regular\_plural\_subject\_verb\_agr\_1         & \textit{Jeffrey hasn’t/*haven't criticized Donald.}                                                                                                              & 99.3                                                                               &                                                                                      &                                                                                     \\
                                                 & regular\_plural\_subject\_verb\_agr\_2         & \textit{The dress/*dresses crumples.}                                                                                                                            & 99.1                                                                               &                                                                                      &                                                                                     \\ 
\hdashline[1pt/1pt]
\multirow{2}{*}{\textsc{npi-only}}               & only\_npi\_licensor\_present                         & \textit{Only/*Even Bill would ever complain.}                                                                                                                    & 100                                                                                & \multirow{2}{*}{0.09}                                                                & \multirow{2}{*}{99.93}                                                              \\
                                                 & only\_npi\_scope                                     & \begin{tabular}[c]{@{}l@{}}\textit{Only those doctors who Karla respects ever ... /}\\\textit{*Those doctors who only Karla respects ever ...}\end{tabular}      & 100                                                                                &                                                                                      &                                                                                     \\ 
\hdashline[1pt/1pt]
\multirow{2}{*}{\textsc{npi-sent-neg}}           & sentential\_negation\_npi\_licensor\_present         & \textit{Those banks had not/*really ever lied.}                                                                                                                  & 100                                                                                & \multirow{2}{*}{0.45}                                                                & \multirow{2}{*}{99.82}                                                              \\
                                                 & sentential\_negation\_npi\_scope                     & \begin{tabular}[c]{@{}l@{}}\textit{The turtles that are boring me could not ever ... /}\\\textit{*The turtles that are not boring me could ever ...}\end{tabular} & \multicolumn{1}{c;{1pt/1pt}}{100}                                                  &                                                                                      &                                                                                     \\ 
\hdashline[1pt/1pt]
\textsc{npi-sim-ques}                            & matrix\_question\_npi\_licensor\_present             & \textit{Should I ever join? / *I should ever join.}                                                                                                              & 100                                                                                & 0.01                                                                                 & 99.98                                                                               \\ 
\hdashline[1pt/1pt]
\multirow{2}{*}{\textsc{quantifier-superlative}} & superlative\_quantifiers\_1                          & \textit{No man has revealed more than/*at least 5 forks.}                                                                                                        & 98.5                                                                               & \multirow{2}{*}{7.29}                                                                & \multirow{2}{*}{97.72}                                                              \\
                                                 & superlative\_quantifiers\_2                          & \textit{An/*No actor arrived at at most 6 lakes.}                                                                                                                & 99.3                                                                               &                                                                                      &                                                                                     \\ 
\hdashline[1pt/1pt]
\textsc{quantifier-existential-there}            & existential\_there\_quantifiers\_1                   & \textit{There aren’t many/*all lights darkening.}                                                                                                                & 99.1                                                                               & 1.15                                                                                 & 99.82                                                                               \\ 
\hdashline[1pt/1pt]
\textsc{binding-c-command}                       & principle\_A\_c\_command                             & \begin{tabular}[c]{@{}l@{}}\textit{A lot of actresses that thought about Alice}\\\textit{healed themselves/*herself.}\end{tabular}                               & 96.6                                                                               & 0.01                                                                                 & 100.0                                                                              \\ 
\hdashline[1pt/1pt]
\multirow{2}{*}{\textsc{binding-case}}           & principle\_A\_case\_1                                & \textit{Tara thinks that she/*herself sounded like Wayne.}                                                                                                       & 100                                                                                & \multirow{2}{*}{1.54}                                                                & \multirow{2}{*}{99.54}                                                              \\
                                                 & principle\_A\_case\_2                                & \textit{Anna imagines herself praising/*praises this boy.}                                                                                                       & 92.5                                                                               &                                                                                      &                                                                                     \\ 
\hdashline[1pt/1pt]
\multirow{3}{*}{\textsc{binding-domain}}         & principle\_A\_domain\_1                              & \textit{Carlos said that Lori helped him/*himself.}                                                                                                              & 100                                                                                & \multirow{3}{*}{0.44}                                                                & \multirow{3}{*}{99.84}                                                              \\
                                                 & principle\_A\_domain\_2                              & \textit{Mark imagines Erin might admire herself/*himself.}                                                                                                       & 99.3                                                                               &                                                                                      &                                                                                     \\
                                                 & principle\_A\_domain\_3                              & \begin{tabular}[c]{@{}l@{}}\textit{Nancy could say every guy hides himself. /}\\\textit{*Every guy could say Nancy hides himself.}\end{tabular}                  & 99.5                                                                               &                                                                                      &                                                                                     \\ 
\hdashline[1pt/1pt]
\textsc{binding-reconstruction}                  & principle\_A\_reconstruction                         & \textit{It’s herself who Karen criticized / *criticized Karen.}                                                                                                  & 99.1                                                                               & 0.01                                                                                 & 99.99                                                                               \\ 
\hdashline[1pt/1pt]
\multirow{2}{*}{\textsc{passive}}                & passive\_1                                           & \textit{Jeffrey’s sons are insulted/*smiled by Tina.}                                                                                                            & 96.9                                                                               & \multirow{2}{*}{2.67}                                                                & \multirow{2}{*}{99.57}                                                              \\
                                                 & passive\_2                                           & \textit{Most cashiers are disliked/*flirted.}                                                                                                                             & 98.9                                                                               &                                                                                      &                                                                                     \\ 
\hdashline[1pt/1pt]
\multirow{4}{*}{\textsc{det-adj-noun}}           & det\_noun\_agr\_with\_adj\_1            & \textit{Tracy praises those lucky guys/*guy.}                                                                                                                    & 95.6                                                                               & \multirow{4}{*}{1.14}                                                                & \multirow{4}{*}{99.78}                                                              \\
                                                 & det\_noun\_agr\_with\_adj\_2            & \textit{Some actors buy these/*this gray books.}                                                                                                                 & 93.0                                                                               &                                                                                      &                                                                                     \\
                                                 & det\_noun\_agr\_with\_adj\_irregular\_1 & \textit{He shouldn’t criticize this upset child/*children.}                                                                                                      & 92.0                                                                               &                                                                                      &                                                                                     \\
                                                 & det\_noun\_agr\_with\_adj\_irregular\_2 & \textit{That adult has brought that/*those purple octopus.}                                                                                                       & 93.9                                                                               &                                                                                      &                                                                                     \\ 
\hdashline[1pt/1pt]
\multirow{4}{*}{\textsc{det-noun}}               & det\_noun\_agr\_1                       & \textit{Craig explored that grocery store/*stores.}                                                                                                              & 99.7                                                                               & \multirow{4}{*}{0.47}                                                                & \multirow{4}{*}{99.95}                                                              \\
                                                 & det\_noun\_agr\_2                       & \textit{Carl cures those/*that horses.}                                                                                                                          & 99.8                                                                               &                                                                                      &                                                                                     \\
                                                 & det\_noun\_agr\_irregular\_1            & \textit{Phillip was lifting this mouse/*mice.}                                                                                                                   & 100                                                                                &                                                                                      &                                                                                     \\
                                                 & det\_noun\_agr\_irregular\_2            & \textit{Those ladies walk through those/*that oases.}                                                                                                            & 100                                                                                &                                                                                      &                                                                                     \\
\arrayrulecolor{black}\bottomrule
\end{tabular}
        }}
        \caption[Training corpora]{ 
        An overview of all the filters, the BLiMP benchmark they target, an example for each benchmark, and number of items targeted by the filter.
        The rightmost column represents the relative number of tokens in each filtered corpus after they have been downsampled to the same number of lines.
        }\label{tab:corpora-stats}
        \end{table*}

    This section first introduces the logic of the FiCT method before detailing the specific filters that we use in our experiments.  The final experimental setup is described in \autoref{sec:setup}. Code and data, as well as a link to all models on the HuggingFace Hub, can be found at \anontext{\url{https://github.com/CLMBRs/corpus-filtering}}.

    \subsection{Logic of the Method}
    \label{sec:fict}
    
    The core methodological basis of this paper is what we call \textit{Filtered Corpus Training}, or FiCT. 
    This involves comparing the performance of otherwise identical learners that are trained on data which differs in some interesting way. 
    
    In this paper, the FiCT methodology is primarily used to test whether LMs are capable of extrapolating linguistic rules learned from environments in training data to unseen environments.
    In order to ensure that the specified environments are not seen in the training data, we use \emph{filters} to remove sentences with the specified environments from a naturalistic corpus.
    By comparing models trained on the ablated data and models trained on the full, naturalistic corpus, we can potentially determine whether, how, and when language models are able to make such generalizations.

    \autoref{fig:overview} illustrates the logic of our method.  The sentence pair ``A sketch of lights \{doesn't / *don't\} appear'' contains a subject with a prepositional phrase (PP) modifying a noun, itself with a noun that differs in number from the main subject.  We filter from the training corpus all sentences with subjects containing PP modifiers, and then compare the ability to make the correct grammaticality judgments on this pair between a model trained on the full corpus and this filtered corpus.  This difference in performance we call $\textsf{acc}\Delta$ (formally defined in \autoref{sec:setup}).  A model that has not seen PP-modified subjects could still make the correct judgments by forming the following generalizations: verbs agree with the head noun of the subject, and noun phrases with PP modifiers (which can be seen in object, but not subject position) are headed by the main noun.  Low $\textsf{acc}\Delta$ would then provide evidence that the model has developed such generalizations.
    
    The filters used in the present investigation are listed in \autoref{tab:corpora-stats}, along with the BliMP benchmark(s) each targets, and some descriptive summary statistics for each.
    These filters utilized part-of-speech, morphological features, and syntactic dependency annotations generated via the use of Stanza \citep{stanza}, an off-the-shelf package that uses pretrained neural models to generate grammatical annotations within the framework of Universal Dependencies (UD) \citep{ud-v1, ud-v2}. We now describe the filters in more detail.

        \subsection{Corpus Filters}
        \label{subsec:filt-impl}

        In general, we favor ``stronger'' filters, i.e.\ those that include false positives (and so filter out more training data), since our goal is to ensure that the LM has not seen a given construction during training. In what follows, $x >_z y$ means that there is a dependency from $x$ to $y$ with label $z$.

            \subsubsection{Structural Generalization}
            
            In the following filters, a particular structural configuration has been completely removed from the corpus, and a model must generalize to it from similar/related configurations.

            \paragraph{\textsc{agr-pp-mod}}

            The benchmark targeted by this filter tests subject-verb number agreement in the presence of an intervening \textit{distractor} in a prepositional phrase, as illustrated in \autoref{fig:overview}. \textsc{agr-pp-mod} filters all sentences containing the dependency structure
            $\textsc{verb} >_\text{nsubj} \textsc{noun} >_\text{nmod} \textsc{noun} >_\text{case} \textsc{adp}$.  The resulting filtered corpus will still contain PPs modifying nouns in other contexts (e.g. object position).  If a learner has formed a general `rule' for subject-verb agreement, and seen PP-modified objects, it should be able to generalize to agreement with PP-modified subjects, even when it hasn't seen them during training.
            
            \paragraph{\textsc{agr-rel-cl}} This filter is similar to the previous one, but targets sentences where the distractor occurs in a relative clause in subject position, removing all sentences containing the structure $\textsc{verb} >_\text{nsubj} \textsc{noun} >_\text{acl:relcl} \textsc{adj}$, e.g.\ ``The boys that aren't disturbing Natalie dream''. 
            A model might generalize again from its general `rule' for subject-verb agreement, and learn about relative clause structure from relative clauses in object position. 

            \paragraph{\textsc{npi-} filters}
            We use the list of negative polarity items (NPIs) provided by \citet{lms-use-monotonicity} and filter as follows: \textsc{npi-only} removes all sentences with an NPI occurring after `only' (e.g. ``Only students have ever complained about morning classes''), \textsc{npi-sent-neg} removes sentences with a negation and an NPI, and \textsc{npi-sim-ques} removes questions with NPIs in them.
            In each of these cases the model can generalize NPI licensing conditions for a particular environment from other environments that are still present.
            
            \paragraph{\textsc{quantifier-superlative}}
            Superlative quantifiers (e.g., \textit{at least}, \textit{at most}) cannot be embedded under negation: \textit{An actor arrived at at most six lakes} vs. *\textit{No actor arrived at at most six lakes}.
            BLiMP targets this phenomenon in two ways: either by replacing the superlative quantifier under negation with a relative quantifier (e.g. \textit{more than 5}), or by removing the negation.
            We cannot detect superlative quantifiers based on dependency information alone, so we use morphological feature annotations.
            Next, we filter all such constructions that appear in object position: $\textsc{verb} >_\text{obl/obj/iobj} \textsc{noun} > \cdots > \textsc{quantifier}$.
            It is less clear for this filter how a model can still infer the grammaticality from other constructions that are not covered by the filter.
            
            \paragraph{\textsc{quantifier-existential-there}}
            \textit{Weak} quantifiers can occur in the scope of existential \textit{there} constructions, whereas \textit{strong} quantifiers can not: \textit{There are many people here} vs. *\textit{There are all people here} \citep{milsark-dissertation-existential-sents}.
            BLiMP targets this phenomenon in two ways: either replacing a weak quantifier with a strong one, or increasing the scope of a locative \textit{there} such that it becomes existential.
            We filter all weak quantifiers occurring in subject position under an existential \textit{there}: $\textsc{there} <_{\text{expl}} \textsc{are} >_\text{nsubj} \textsc{noun} > \textsc{weak-Q}$.
            However, we only filter the 5 weak quantifiers occurring in the BLiMP benchmark (\textit{a(n), no, some, few, many}), which still allows a model to generalize from other weak quantifiers to infer the grammaticality conditions.
            Furthermore, weak vs. strong quantification plays a role in other linguistic phenomena as well, a fact which a learner could leverage.

            \paragraph{\textsc{binding-} filters}
            Four filters, \textsc{binding-c-command}, \textsc{binding-case}, \textsc{binding-domain}, and \textsc{binding-}\textsc{reconstruction} target the seven binding-related benchmarks of BLiMP. All seven benchmarks typify various facets of \citet{chomsky1993-gov+binding}'s Principle A. 
            The implementations of all four filters is generally similar: they target sentences where a reflexive or non-reflexive pronoun occurs in the specific context(s) illustrated by the corresponding benchmarks, narrowly construed, while leaving in sentences where the same or similar principle is applied in a different environment. 
            For example, the \textsc{binding-c-command} filter removes evidence of the use of the c-command relationship in anaphora licensing \textit{in relative clauses}, but not elsewhere, as in sentences like \textit{Mary's brother hurt himself} (but not *\textit{Mary's brother hurt herself}).\footnote{BLiMP assumes a straightforward one-to-one relationship between certain names and their grammatical gender. While such a relationship may not actually be borne out in practice today, the corpora used in this investigation likely do adhere to such a formulation.} The other three benchmarks operate in similar ways.

            \paragraph{\textsc{det-adj-noun}}
            One of the filters targeting determiner-noun agreement focuses on cases where an adjective occurs between a demonstrative determiner and a noun, e.g. \textit{These/*This red cars}.
            We create a filter that removes \textit{all} occurrences of a demonstrative determiner followed by an adjective and a noun.
            A model can then still infer the number agreement from determiner/noun pairs without an intervening adjective.

            \subsubsection{Lexical Generalization}
            In the following filters we do not filter out an entire configuration, but only do so for a subset of lexical items. 
            This way a model can indirectly generalize to a specific occurrence of the configuration from other occurrences, but no longer rely on direct co-occurrences. These filters focus on lexical generalization because the BLiMP benchmarks that they target are centered around particular lexical items and not particular syntactic constructions.
            
            \paragraph{\textsc{agr-re-irr-sv}}
            The four BliMP benchmarks targeted by \textsc{agr-re-irr-sv} all test language model performance on subject-verb agreement, targeting regular plurals, like \textit{dress}/\textit{dresses} and irregular plurals, like \textit{goose}/\textit{geese}. The filter removes all sentences with nominal subjects where the noun occurs in any of the four benchmarks. A learner on this filtered corpus can still \new{beat} the benchmark if it develops a notion of grammatical number, a representation of the grammatical number of the nouns in the benchmark based on their usage in other contexts, and then generalizes the subject-verb agreement it sees for other nouns to these nouns.
           
            \paragraph{\textsc{det-noun}}
            The other filter besides \textsc{det-adj-noun} that targets determiner-noun agreement for demonstrative determiners (e.g. \textit{These}/*\textit{This books}) does so with the determiner directly adjacent to the noun.
            We create a filter based on all nouns occurring in the BLiMP benchmark that are preceded by a demonstrative determiner.
            A model can still infer the number agreement between determiner and noun from other nouns, and learn the number information of the filtered nouns from other agreement tasks like subject-verb agreement.

            \paragraph{\textsc{passive}}
            \new{In English,} passive constructions can only be formed from transitive verbs.
            BLiMP targets this phenomenon by replacing transitive verbs in passive constructions by intransitive verbs: \textit{John is insulted by Mary} vs. *\textit{John is smiled by Mary}.
            Much like \textsc{agr-re-irr-sv} and \textsc{det-noun}, the \textsc{passive} filter operates by removing sentences that contain words on a word list in a specific linguistic environment. 
            Concretely, this word list consists of the verbs that are actually used in these two benchmarks in passive form, and the filter removes sentences where such words appear in passive voice.

    \section{Experimental Setup}
    \label{sec:setup}
        \paragraph{Data}\label{sec:data}
        The base train, validation, and test 
        corpora are the English Wikipedia corpora released by \citet{gulordava2018}, with the train corpus consisting of 3.05M sentences (83M tokens, with a vocabulary size of 50000 plus an unknown and EOS token). 
        The 15 filtered corpora are derived from this base corpus by discarding all sentences that are targeted by the filter.
        The number of sentences and tokens discarded by each filter varied from as little as ${\sim}0.1\%$ to as much as ${\sim}18.5\%$; for specifics, refer to \autoref{tab:corpora-stats}. 
        Then, as an additional control, the fifteen filtered corpora plus the original, \textsc{full} training corpus were uniformly downsampled to 2.4M lines, corresponding to ${\sim}80\%$ the size of the original training corpus. 
        It is worth noting that the number of \textit{tokens} did vary by as much as ${\sim}4.2\%$, as reflected in the rightmost column of \autoref{tab:corpora-stats}: this is explained by the fact that certain filters target longer sentences more often.

        \paragraph{Models}
        Two architectures are used for the models trained in this investigation: LSTMs \citep{hochreiter1997-lstm}) and decoder-only Transformers \citep{attention-is-all-you-need}. 
        For each architecture, we train separate models on the 16 training corpora for five random seeds each, resulting in a total of 160 models.        
        Model hyperparameters were selected to control for number of parameters as closely as possible. 
        The LSTMs have two layers with embedding and hidden dimension of 1024. 
        Output and embedding layer weights were tied, and we used dropout of 0.1 during training. 
        The Transformers were constructed with feed-forward and hidden layer dimensions of 768, eight attention heads, and eight hidden layers.
        The LSTMs and the Transformers had 68.0M and 67.1M trainable parameters, respectively.

    \paragraph{Training}
    Each model was trained on a single A40 GPU for 40 epochs with mixed-precision training, using the AdamW optimization algorithm \citep{loshchilov2017-adamw}, a linear scheduler with an initial learning rate of \num{5e-05}, and a batch size of 32.  
    We evaluated each model at the end of every epoch, and report results for the model with the best validation perplexity. The full hyperparameter set may be found in \autoref{app:hyperparams}.
  
    \paragraph{Evaluation}
    \label{sec:eval}
    We use four metrics---three standard and one novel---as the primary means of evaluation for all models. 
    The first is perplexity over the (unfiltered) test corpus of \citet{gulordava2018}. 
    The second is accuracy on each of the 67 benchmarks in the BLiMP challenge set \citep{blimp}. Accuracy on the BLiMP benchmarks was assessed via the ``full-sentence'' method \citep{marvin2018-targeted}, where a ``success'', for any minimal pair, is defined by the model assigning a higher probability to the grammatical sentence in the minimal pair ($s^+$) than to the ungrammatical sentence ($s^-$). 

    However, the FICT methodology's main advantage lies not in looking at the performance of each model in isolation, but on the \textit{difference} in performance between two models that are otherwise identical but for their training data. Thus, for each model and each BLiMP benchmark, a change score (or delta) was calculated with respect to the average performance of all models of the same architecture trained on the \textsc{full} corpus (i.e. average over the five seeds).  
    
    To be more precise, with $M$ a model type (i.e.\ $M \in \{ \textsf{LSTM}, \textsf{Transformer} \}$), $F$ a filter, and $B$ a benchmark, $F(B)$ will refer to the filtered corpus targeting $B$, and $M_F$ will refer to a model trained on $F$. We can then define the accuracy delta by:
    \begin{equation}
        \textsf{acc}\Delta(M, F, B) := \textsf{acc}^{M_F}_B - \overline{\textsf{acc}^{M_\textsc{full}}_B}
    \end{equation}
    where $\textsf{acc}^M_B$ refers to the accuracy of model $M$ on benchmark $B$.
    We will often be interested in the case where $F = F(B)$, i.e. the benchmark(s) corresponding to the corpus filter, but report others as well.

    Our final evaluation metric looks at the \textit{probability deltas} between grammatical and ungrammatical sentences: 
    \begin{equation}
    P\Delta(M, F)(s) = \log P_{M_F}(s^+) - \log P_{M_F}(s^-)
    \end{equation}
    \new{$P\Delta$ expresses the magnitude of a model's grammaticality judgment: whereas $\textsf{acc}\Delta$ only expresses the ratio of items for which a model assigned a higher probability to the grammatical case, $P\Delta$ can be interpreted as the confidence of a model's judgment.}
    
\section{Results}
\label{sec:results}
    We present our results along the four metrics of \S\ref{sec:setup}: perplexity (\S\ref{sec:results/ppl}), TSE accuracy (\S\ref{sec:results/blimp-acc}), accuracy delta (\S\ref{sec:results/acc-delta}), and probability delta (\S\ref{sec:results/prob-delta}).

    \begin{figure*}[ht]
        \centering
        \includegraphics[width=\textwidth,clip,trim={0.5cm 0 0.9cm 0}]{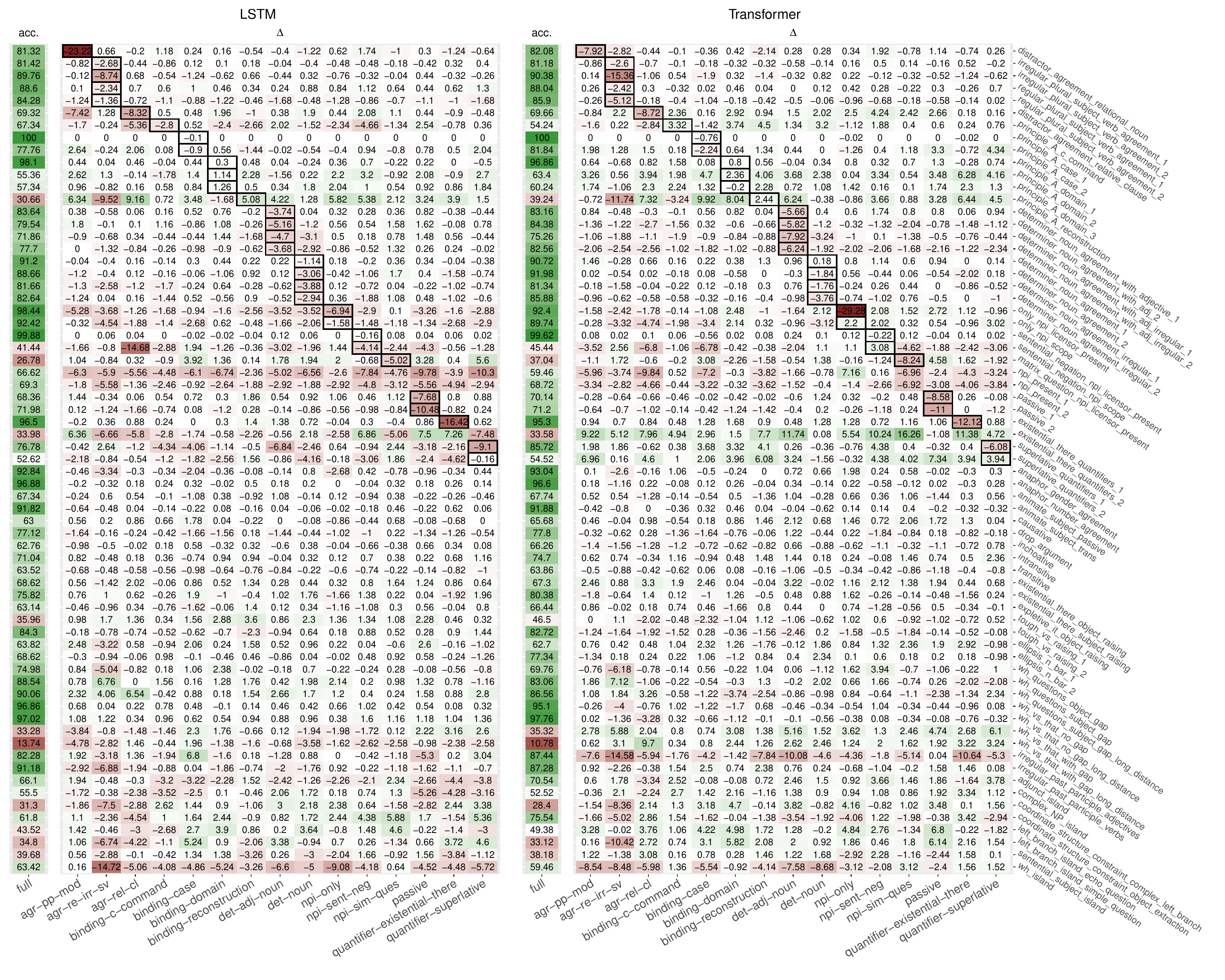}
        \caption[]{
        BLiMP benchmark \textit{accuracy} for the models trained on the full corpus, and \textit{accuracy delta} ($\Delta(M, F, B)$) for the filtered corpora, averaged across seeds. 
        Boxes with bold outlines correspond to benchmarks targeted by the model's corpus filter (i.e.\ where $F = F(B)$). 
        The accuracy scored by a given model on a given benchmark trained on a filtered corpus can be recovered by adding its delta to the accuracy score in the ``full'' column of the same row.
        }
        \label{fig:blimp-delta-heatmap}
    \end{figure*}

    \subsection{Perplexity}
    \label{sec:results/ppl}
    
        We found that Transformers uniformly achieve lower perplexities on the test corpus than the LSTMs for all training corpora, as expected.  The mean test perplexity across all corpora and random seeds was $47.13$ for the Transformers and $53.56$ for the LSTMs; a paired $t$-test of mean perplexities per corpus found the difference between the model types to be significant ($t = 270.94$, $p \ll 0.01$).
        As noted in \autoref{sec:data}, while we downsampled all corpora to the same number of lines, the number of tokens varies between different training corpora. Previous research has shown a clear negative relationship between the number of tokens seen in training and test corpus perplexity \citep{DBLP:journals/corr/abs-2001-08361}. 
        This effect is also present in our data, for both architectures (LSTMs: Pearson's $r=-0.970$; Transformers: $r=-0.976$).

\begin{figure}
        \centering
        \includegraphics[width=2.5in]{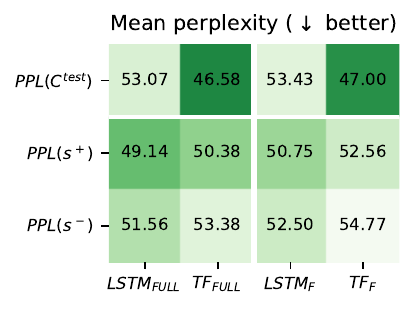}
        \caption{Perplexity scores on the test corpus ($C^{\textit{test}}$) and the grammatical and ungrammatical BLiMP sentences ($s^+$ \& $s^-$).
        BLiMP scores for the \textsc{full} models are averaged over all benchmarks, and for the Filtered models for their corresponding benchmark.
        }
        \label{fig:blimp-ppl}
    \end{figure}

        \new{
        We also investigate the perplexity on the BLiMP sentences for the \textsc{full} and Filtered models.
        This provides us insight into the likelihood of these sentences: if the model assigns a relatively low likelihood to them, then grammaticality judgments will be less reliable as well \citep{newman-etal-2021-refining}.
        In Figure~\ref{fig:blimp-ppl} we show the scores for this.
        Surprisingly, the LSTM models yield \textit{lower} perplexity on the BLiMP sentences than the Transformers.
        This shows that Transformers have shifted their probability mass to other sentence types than found in BLiMP, but where to exactly remains an open question.
        Nonetheless, the perplexity scores on BLiMP are similar to the average perplexity on the test corpus, which demonstrates that these items are of similar likelihood. 
        } 

    \subsection{TSE Accuracy on BLiMP}
    \label{sec:results/blimp-acc}

    Mean overall accuracy on all of BLiMP across different training corpora (i.e.\ $\overline{\textsf{acc}_\textsc{all}^{M_F}}$) was $70.4$ for the LSTMs and $71.9$ for the Transformers.  This result was statistically significant (paired $t=-17.38$, $p \ll 0.01$).  \autoref{app:fig:blimp-acc-heatmap} in \autoref{sec:app-blimp-acc} shows all of the accuracies.

    We next look only at benchmark accuracy data where the filtered corpus targeted a given benchmark, i.e.\ where $F=F_B$.  Here, the mean is $68.8$ for the Transformers and $66.7$ for the LSTMs \textit{and this difference is not statistically significant} (paired $t = -1.18$, $p=0.258$).  In other words, we find no difference in the two models' ability to make grammaticality judgments when trained on filtered data that forces them to perform subtle generalizations, despite differences in perplexity.

    \subsection{Accuracy Delta}\label{sec:results/acc-delta}

    A table of the accuracy deltas, averaged across all random seeds, can be found in \autoref{fig:blimp-delta-heatmap}. 
    Mean overall accuracy delta over all benchmarks and across all training corpora (i.e.\ $\overline{\Delta}(M, F, B)$) was $-0.393$ for the LSTMs and $0.0313$ for the Transformers.  This result was statistically significant (paired $t=-5.10$, $p \ll 0.01$).

    Focusing on the $F = F(B)$ cases (i.e.\ black-outlined cells in the chart), we note that most deltas are generally negative but fairly close to zero, with a few outliers, such as the models trained on the \textsc{existential-there}, \textsc{agr-pp-mod}, and \textsc{npi-only} corpora. These results suggest that, overall, learners \textit{are} usually able to use various sorts of indirect evidence to acquire correct grammatical generalizations when direct evidence has been made unavailable, as otherwise we could expect much larger deltas across the board.
    
    We may also observe that, for the cases where the absolute value of the deltas was appreciably larger than zero, it is not the case that one architecture is uniformly better than the other. For example, LSTMs perform better than Transformers (that is, their deltas are smaller in magnitude) on the benchmarks associated with the \textsc{agr-re-irr-sv} and the \textsc{npi-only} corpora, while the converse is true for \textsc{agr-pp-mod} and quantifier-existential-there. This is true \textit{even} for phenomena that are seemingly relatively similar; for example, the \textsc{agr-pp-mod} and \textsc{agr-re-irr-sv-agr} filters are extremely similar, in that they both test long distance agreement in the present of a clausal distractor intervening between the subject and the verb; they differ only in the nature of that distractor.
    Yet, as noted, LSTMs trained on the \textsc{agr-re-irr-sv} corpus have, on average, a less negative delta on the associated benchmarks than the analogous Transformer models ($\overline{\textsf{acc}\Delta}(\text{LSTM}, \textsc{agr-re-irr-sv}, F(B)) = -3.78$; for the Transformer, $-6.38$); conversely, on the models trained on the \textsc{agr-pp-mod} corpus, it is Transformers which have the smaller magnitude delta ($\overline{\textsf{acc}\Delta}(\text{LSTM}, \textsc{agr-pp-mod}, F(B)) = -23.22$; Transformer, $-7.92$). 
    
    As in the previous section, we can make this precise by analyzing all of the accuracy deltas where $F=F_B$. 
    The mean here is $-5.41$ for the LSTMs and $-4.62$ for the Transformers and this difference is not statistically significant (paired $t = -0.562$, $p=0.583$). 
    That means that we again find no difference between the two architectures in the extent to which filtering affects their accuracy, despite significant differences in perplexity. 
    This suggests that perplexity \textit{does not} predict the ability of a model to perform linguistic generalizations from indirect evidence.

    \begin{figure*}[t!]
        \centering
        \includegraphics[width=0.65\columnwidth]{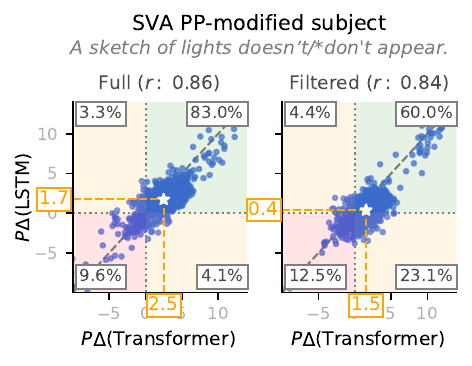}\quad
        \includegraphics[width=0.65\columnwidth]{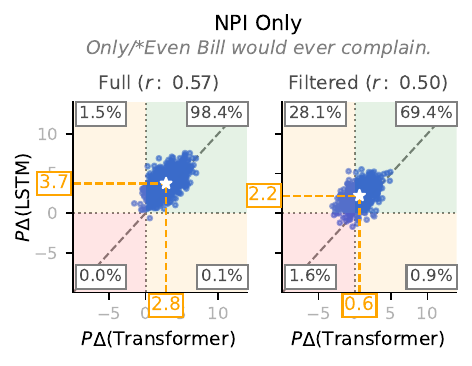}\quad
        \includegraphics[width=0.65\columnwidth]{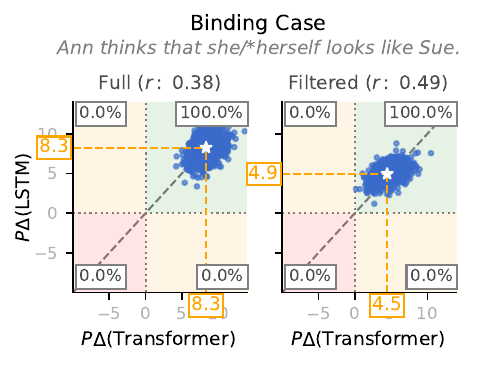}\quad
        \caption{
        Log probability differences between grammatical and ungrammatical minimal pairs ($P\Delta(M, F)(s)$), with Transformer performance plotted against LSTM performance.
        Individual points are the averaged scores across the five model seeds.
        The four quadrants indicate the cases where i) both architectures got a correct prediction (\textcolor{olive}{green}), ii) only one architecture got a correct prediction (\textcolor{orange}{orange}), and iii) neither architecture was right (\textcolor{red}{red}).
        It can be seen that corpus filtering results in probability differences moving closer to the origin, and that the magnitude of the difference of the full models can create a sufficient margin for the model to generalize in the filtered cases as well.
        }
        \label{fig:prob-delta-scatter}
    \end{figure*}

    \begin{figure}[t]
        \centering
        \includegraphics[width=\columnwidth]{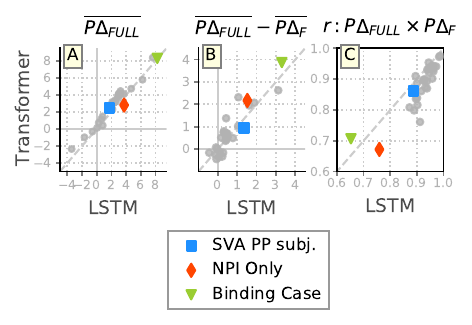}
        \caption{
        A: $P\Delta$ scores for the \textsc{full} Transformers and LSTMs for each BLiMP paradigm.
        The more positive this score, the more certain a model is in its grammaticality judgment.
        B: Paradigm-level differences in $P\Delta$ scores going from the \textsc{full} to the Filtered model. 
        The closer to the origin, the less impact the filtering procedure had on model behavior.
        C: Pearson correlation of $P\Delta$ scores between the \textsc{full} and Filtered models.
        A detailed table with these results per paradigm is provided in Figure~\ref{app:fig:pdelta-table} in Appendix~\ref{sec:app-blimp-acc}.
        }
        \label{fig:pdelta-benchmark-level}
    \end{figure}

    \subsection{Probability Delta}\label{sec:results/prob-delta}
    In order to gain a more fine-grained insight into the impact of corpus filtering, we examine the results at an item-level.
    For this we make use of the $P\Delta$ metric, which expresses a model's magnitude of a grammaticality judgment.
    In \autoref{fig:pdelta-benchmark-level}A we plot the average $P\Delta$ scores for the \textsc{full} models for each BLiMP benchmark, averaged across seeds.
    It can be seen here that the Transformers and LSTMs result in highly similar $P\Delta$'s ($r=0.98;p\approx 0$), although the Transformer scores are slightly higher on average \new{than} those of the LSTMs ($2.99$ vs. $2.41$, respectively), which is in line with the significant difference in TSE accuracy of \S\ref{sec:results/blimp-acc}.
    
    For the sake of brevity, we focus on three salient filters that each yielded distinct results: i) Subject-Verb Agreement for PP-modified subjects, in which LSTMs are more impacted than Transformers (${\textsf{acc}\Delta}$: $-23.2$ vs. $-7.9$); ii) NPI Only, in which LSTMs are \textit{less} impacted than Transformers (${\textsf{acc}\Delta}$: $-6.9$ vs. $-29.3$); and iii) Binding Case, in which neither architecture is impacted by filtering.
    In \autoref{fig:prob-delta-scatter} we plot the item-level scores of the LSTMs against the Transformers (averaged across seeds).
    For each benchmark $B$ we plot the results on the \textsc{full} model and the $F(B)$ filtered model.
    This demonstrates that corpus filtering has the effect of moving $P\Delta$ closer to the origin: the model becomes \textit{less certain} in its grammaticality judgment.
    The resulting $\textsf{acc}\Delta$ score for a benchmark is then dependent on the $P\Delta$ scores of the \textsc{full} model: a sufficient margin here makes it robust to the decrease in $P\Delta$ and allows it to correctly assign higher probability to the grammatical item.

    To investigate this observation across all benchmarks we plot the difference in $P\Delta$ going from \textsc{full} to Filtered in \autoref{fig:pdelta-benchmark-level}B.
    This difference represents the \textit{absolute impact} of filtering on the TSE task.
    By plotting the Transformer results against the LSTM we gain an insight whether filtering has a similar impact on both architectures.
    We observe a strong correlation between these $P\Delta$ differences ($r: 0.91$, $p\approx 0$).
    Subtle difference are present, however, for a number of filters the $P\Delta$ score \textit{increases} after filtering which is especially prevalent for the Transformer models.

    Finally, we examine the \textit{robustness} of a model's grammaticality judgments: does filtering have a significant impact on the distribution of judgments?
    For this we compute the Pearson correlation of $P\Delta$ before and after filtering for each filter benchmark.
    A model is robust to filtering if this correlation remains high.
    In \autoref{fig:pdelta-benchmark-level}C we plot the LSTM correlations against the Transformer.
    A striking difference between the two architectures arises here: the LSTM correlations are systematically larger than those of the Transformer.
    This shows that LSTMs are less impacted by filtering on an item-level than Transformers.


\section{Discussion}
\label{sec:discussion}


\paragraph{Perplexity Versus Linguistic Generalization}
Our findings contribute to a growing body of research that suggest a dissociation between perplexity and more targeted evaluations of linguistic competence in artificial learners \citep{hu2020-systematic}. 
In a carefully controlled setting and for a wide range of phenomena, we demonstrate that the training objective of minimizing perplexity does not predict linguistic generalization.
This raises interesting questions on the relation between perplexity and grammaticality judgments \citep{Lau2017GrammaticalityAA}: while Transformers are better at \textit{memorizing} the structure of its training data, we show they are less capable than LSTMs of forming robust linguistic generalizations.
An interesting step for future work would be to uncover what language modeling aspects Transformers \textit{do} excel at, which allows them to obtain a superior test perplexity \new{(e.g. word frequency, as studied in \citealp{wei-etal-2021-frequency})}.  
\new{Future work should also compare our measure(s) of generalization with others in the literature, given evidence that these are not always well-correlated with each other \citep{sun-etal-2023-validity}.}

\new{We also note that while likelihood judgments do not necessarily directly measure grammaticality, since likelihood is the outcome of many other factors (e.g. semantic plausibility, pragmatic felicity), the use of minimal pairs for BLiMP does help control for this since it reports judgments on sentences which differ on (usually) one word, thus keeping these other components constant between the two sentences.  That being said, it would be a worthwhile follow-up to conduct probing experiments to more directly model grammaticality judgments, in the style of e.g. \citet{lms-use-monotonicity} (see the next subsection as well).\footnote{We thank an anonymous reviewer for encouraging us to think about this distinction.}}

\new{Our results also have consequences for how we think about language model evaluation more broadly: to the extent that we believe that models should be able to generalize from indirect evidence, we cannot rely on perplexity as the sole measure of LM quality but must measure and test for this ability directly.}

\paragraph{Generalizing from Indirect Evidence}
Our study also builds on the insights of numerous other works that use artificial learners as models for understanding human language acquisition, and gaining better insights in the inductive biases of such learners \citep{warstadt2020bowman,mueller-linzen-2023-plant,weber2024}. 
The present study conducts for a wide range of phenomena what \citet{warstadt-dissertation} calls a ``proof-of-concept [of a] large-scale controlled ablation study on the input to model learners,'' and finds that direct attestation of linguistic evidence is not strictly necessary for the development of sophisticated linguistic generalizations. 
Rather, learners can leverage much more indirect sources of evidence to arrive at the correct generalizations.

Where earlier work has focused on specific linguistic constructions, such as subject auxiliary inversion \citep{warstadt-dissertation}, relative clauses \citep{prasad-etal-2019-using}, and negative polarity items \citep{warstadt-etal-2019-investigating, lms-use-monotonicity, weber-etal-2021-language}, the results of this paper essentially confirm a similar result for a much wider array of syntactic and semantic phenomena. 
While in many cases the ablations we performed did clearly negatively affect the performance of our artificial learners on the relevant linguistic evaluations, the magnitude of this effect was generally quite small for all but a small handful of the linguistic phenomena we analyzed. 
In general, even when tested on the specific benchmarks corresponding to the environments that were ablated from their input, models still perform considerably better than chance. 
Thus, our research provides evidence in favor of the indirect evidence hypothesis.

Notably, we find that this is true not only for filters where there are fairly obvious sources of indirect evidence (as enumerated in \autoref{sec:filters}), but also for filters where potential sources of indirect evidence for a correct generalization are much less clear (such as the \textsc{superlative-quantifier} filter). 
This suggests that there may be complex mechanisms by which certain linguistic generalizations can be derived via highly indirect means.
Thus, our results open a door to future research that can provide a more thorough account of the source of these generalizations, with potentially significant ramifications for linguistics.

\paragraph{Explaining Linguistic Generalization}

As just discussed, the primary contribution of this paper has been the development of the FiCT method and the use of it to demonstrate LMs' successful generalization from indirect evidence across a \textit{wide range} of linguistic phenomena.  This success raises a very natural follow-up question: what explains this successful generalization behavior?

While a complete answer to this question must await future work, a detailed look at the NPI cases can provide insight into what an answer may look like.  \citet{lms-use-monotonicity} used a filtered corpus method to test LSTM LMs' understanding of negative polarity items, but then also did a further analysis to examine the basis upon which the models made their grammaticality judgments.  In particular: they found (via probing classifiers) that LMs' were successfully recognizing the \textit{monotonicity} of a linguistic environment and (via a novel correlation method) that these judgments of monotonicity were highly correlated with the LMs' judgment of NPI acceptability, reflecting human acceptability judgments \citep{denicInfluencePolarityItems2021, chemlaModularityIntuitionsFormal2011}.

This example suggests two paths forward for explaining the generalization observations in the present paper. On the one hand, in the same way that the monotonicity explanation was inspired by human generalization, detailed explanations of individual cases of generalization can be developed with human behavior as an initial inspiration.  On the other hand, in the same way that this paper extends the filtered corpus training method to a much wider range of phenomena, one can attempt to generalize these forms of explanation on the breadth axis as well.  We leave these exciting pursuits to future work.

\section{Conclusion}

We introduced the \textbf{Fi}ltered \textbf{C}orpus \textbf{T}raining methodology and applied it to a wide range of linguistic constructions from the BLiMP benchmark.  Our results show that while Transformers are better language models (via perplexity) than comparable LSTMs, the latter generalize equally well (via $\textsf{acc}\Delta$ and $P\Delta$).  The relatively low $\textsf{acc}\Delta$ scores in general show that all of our LMs exhibit a strong ability to generalize from indirect evidence, even for models of relatively low parameter count trained on relatively small data.
\new{In summary, this shows that language model success cannot be attributed solely to memorization from its training data, since the data has been systematically purged of the evaluation targets.  They are, instead, able to form subtle and linguistically-relevant generalizations from indirect evidence.}

Future work will (i) extend this approach to models of different sizes and pretraining corpora, (ii) perform deeper analyses of the bases on which the models do make their generalizations \new{(including with probing experiments)}, and (iii) analyze other forms of lexical and structural generalization through the lens of the filtered corpus training methodology.

\iftaclpubformat
\section*{Acknowledgements}
For helpful discussion, we thank Milica Deni\'c, Dieuwke Hupkes, Jakub Szymanik and the audience at the UW Computational Linguistics Treehouse.

\section*{Author Contribution statement}
Following a practice in several other fields, we here list author contributions according to the Contributor Role Taxonomy (CRediT; \citealp{allenHowCanWe2019}).
\textbf{Abhinav Patil}: Conceptualization, Methodology, Software, Formal analysis, Investigation, Data curation, Writing---original draft, Writing---review and editing, Visualization, Supervision.
\textbf{Jaap Jumelet}: Conceptualization, Methodology, Software, Formal analysis, Data curation, Writing---original draft, Writing---review and editing, Visualization, Supervision.
\textbf{Yu Ying Chiu}: Software, Data curation, Writing---review and editing.
\textbf{Andy Lapastora}: Methodology, Software, Investigation, Data curation, Writing---review and editing.
\textbf{Peter Shen}: Software, Data curation, Writing---review and editing.
\textbf{Lexie Wang}: Software, Data curation, Writing---review and editing.
\textbf{Clevis Willrich}: Software, Data curation, Writing---review and editing.
\textbf{Shane Steinert-Threlkeld}: Conceptualization, Methodology, Software, Formal analysis, Resources, Writing---original draft, Writing---review and editing, Supervision, Project administration.
\fi

\bibliography{fict}
\bibliographystyle{acl_natbib}

\clearpage

\appendix
%
%

\section{Training Hyperparameters}
\label{app:hyperparams}
\begin{table}[h]
{\centering
\begin{tabular}{@{}ll@{}}
\toprule
adam\_beta1                  & 0.9         \\
adam\_beta2                  & 0.999       \\
adam\_epsilon                & 1e-08       \\
dataloader\_num\_workers      & \texttt{8}           \\
evaluation\_strategy         & epoch       \\
fp16                        & True        \\
gradient\_accumulation\_steps & 1           \\
ignore\_data\_skip            & True        \\
learning\_rate               & 5e-05       \\
lr\_scheduler\_type           & linear      \\
num\_train\_epochs            & 40          \\
per\_device\_train\_batch\_size & 32          \\
per\_device\_eval\_batch\_size  & 32          \\
optim                       & adamw\_torch \\
seed                        & {0,1,2,3,4} \\
save\_strategy               & epoch       \\ \bottomrule
\end{tabular}
}
\caption[Training hyperparameters]{Selected training hyperparameters, as provided to the \texttt{transformers} package's \texttt{TrainingArguments} class. Any omitted values were set to the defaults associated with version 4.30.2 of the \texttt{transformers} package.}\label{app:tab:hyperparams}
\end{table}

\section{Full Result Tables}
\label{sec:app-blimp-acc}

\begin{figure*}[ht]
        \centering
        \includegraphics[width=\textwidth]{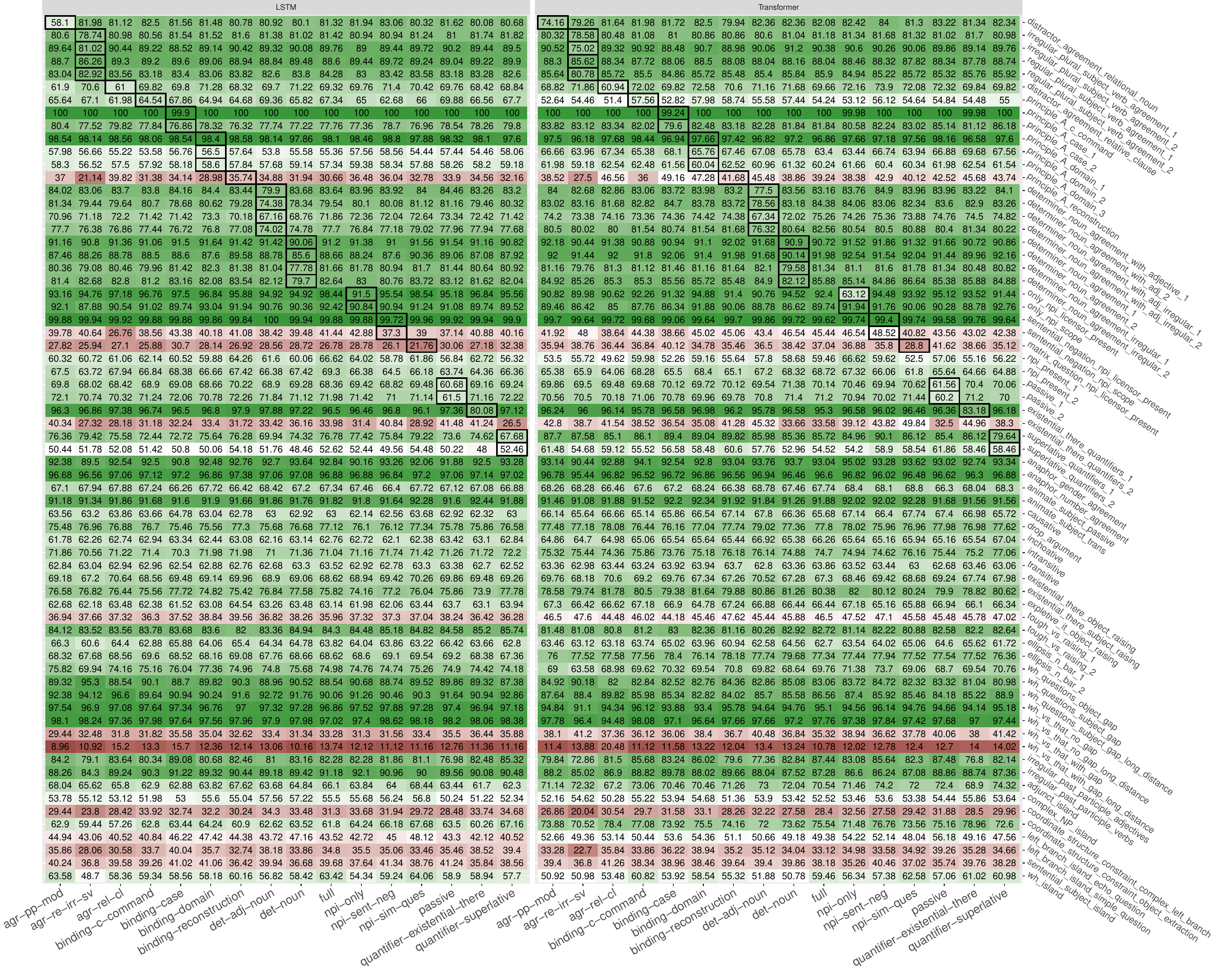}
        \caption[]{
        Complete BLiMP benchmark accuracy results for all models, averaged across the five starting seeds for a given training corpus and benchmark. 
        Boxes with bold outlines correspond to benchmarks targeted by the model's corpus filter (i.e.\ where $F = F(B)$).
        }
        \label{app:fig:blimp-acc-heatmap}
    \end{figure*}

\autoref{app:fig:blimp-acc-heatmap} contains the mean accuracies (across random seeds) on all BLiMP benchmarks for both models and every filtered corpus.

\begin{figure*}[ht]
        \centering
        \includegraphics[width=0.7\textwidth]{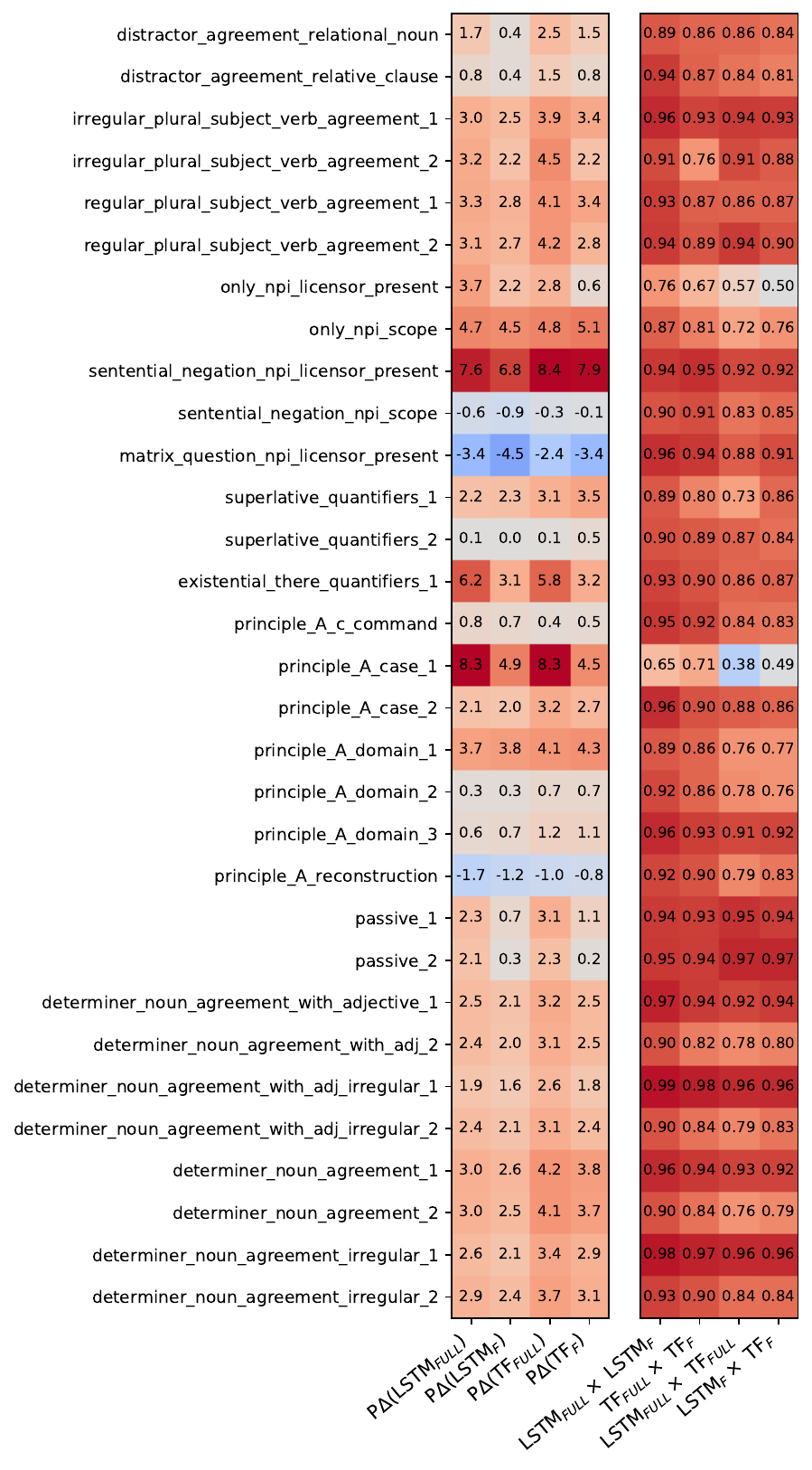}
        \caption[]{$P\Delta$ scores for the LSTMs and Transformers (first four columns), and the Pearson correlations between these $P\Delta$ scores (last four columns).}
        \label{app:fig:pdelta-table}
    \end{figure*}

\autoref{app:fig:pdelta-table} contains the paradigm-level $P\Delta$ scores for the \textsc{full} and Filtered models, and various Pearson correlations.

\end{document}